%% file: main.tex
\documentclass[11pt]{article}
\usepackage[total={5.5in,8in}]{geometry}
\usepackage{tikz}
\usepackage{tabularx}
\usepackage{booktabs}
\usepackage[round]{natbib}

\usepackage{hyperref}

\title{Evaluating RAG Metrics in Applied Contexts: An Experiment, Its Findings and Its Limitations}

\author{Quentin Brabant\\
    Orange Research, Lannion, France\\
    \texttt{quentin.brabant@orange.com}
 }

\begin{document}

\date{}

\maketitle

\begin{abstract}
  This paper reports an empirical study evaluating the relevance of several RAG metrics.
The experiment is based on a question-answering dataset created by human annotators from business data. The generated responses and retrieved spans of a RAG system are scored using evaluation metrics from four libraries (Ragas, DeepEval, RAGChecker, Opik). These metrics are compared to scores given by two evaluators, as well as to standard metrics such as recall. An analysis of correlations is conducted. Finally, we highlight certain limitations of our methodology, compare it to those used in the literature, and suggest some avenues for future research. This paper is an English translation of a paper originally published in the French-speaking workshop EvalLLM 2026 \citep{brabant_2026}.
\end{abstract}

\section{Introduction}

Evaluating and comparing RAG (\emph{Retrieval Augmented Generation}) systems remains a challenging task today: even when a sufficiently large set of test questions with reference answers is available, automatically evaluating a system's responses against these references is far from trivial. A popular approach is to use so-called LLM-as-a-judge metrics to perform this evaluation. Although these metrics generally seem more relevant than classical metrics such as BLEU, it is difficult to know in advance what the relevance of a specific metric will be on the dataset under consideration, especially since evaluation criteria can vary (relevance, factuality, completeness of responses, etc.). It is therefore useful, when evaluating a RAG system under development, to conduct an evaluation of available metrics in order to verify that they provide acceptable approximations of the criterion considered, and that they will thus enable reliable comparison of different iterations of the RAG system being developed. Generally, metrics are evaluated by measuring their correlation with scores given by humans.

This article reports an experiment of this type.
This experiment is based on a question-answering dataset created by annotators from business data. The responses produced and the documents retrieved by a RAG system are scored using RAG metrics from four libraries: Ragas\footnote{\url{https://docs.ragas.io/ }}\citep{es_ragas_2024}, DeepEval\footnote{\url{https://docs.confident-ai.com/}}, RAGChecker\footnote{\url{https://github.com/amazon-science/RAGChecker}}\citep{ru_ragchecker_2024}, and Opik\footnote{\url{https://www.comet.com/site/products/opik/}}. These metrics are compared to reference evaluations: human evaluations for assessing generated responses, and recall for evaluating retrieval.

Note that our objective is not to compare the different metrics evaluated, as the study results are strongly dependent on the choices made during its design. The reported experiments rather aim to apply a methodology in order to test its advantages and limitations.
Unfortunately, the question-answering dataset used cannot be made public; however, we share the raw scores and the code used for the statistical analyses\footnote{\url{https://github.com/Orange-OpenSource/evalllm2026-metric-correlation-analysis}}.

The article is organized as follows.
Section \ref{sec:context} describes the application context and the data used, including manual annotation and evaluation processes. Section \ref{sec:method} describes the methodology applied to compare reference evaluations with evaluations produced by metrics from the tested libraries.
Section \ref{sec:results} reports and analyzes the results of this experiment. Certain limitations of our methodology are highlighted in Section \ref{sec:limites}. In Section \ref{sec:sota}, our methodology is compared to those employed in the literature. We finally propose some research perspectives aimed at facilitating the application of reliable methodologies for evaluating RAG metrics, in Section \ref{sec:pistes}.

\section{Context and Data}\label{sec:context}

Our company is developing a RAG solution designed to answer questions in French about a business domain. The developed system processes each given question via two key modules: first, a \emph{retriever}, whose role is to retrieve relevant spans within a business document database (also in French), which combines a dense approach with BM25; then a \emph{generator}, which injects the user's question and the top 5 spans from the retriever into a prompt for GPT 3.5, so that it generates the answer to the user's question. The retriever is a hybrid system combining a dense approach with BM25.

In order to evaluate the performance of this system, a dataset was created from the corpus of business documents used by the RAG system. This dataset is a set of question-answer pairs accompanied by reference spans.

\subsection{The Question-Answering Dataset}

This dataset consists of 96 questions. Each question is associated with a reference answer, as well as one or more spans from business documents; these spans contain the information necessary to answer the question with which they are associated.

The question-answering dataset is created based on the existing business document database on the telecommunications domain, containing information on, for instance, different offers and customer relations. It contains 479 documents, with length ranging from 11 to 14,025 words (1,112 on average).
The questions and answers are written by annotators who are company employees, as follows:

\begin{enumerate}
    \item Documents are divided into ``pages'' of maximum 1,000 words, relying on the structure of titles and sections to preserve content coherence. This size limit is chosen arbitrarily, with the aim of limiting the amount of information to process for each annotation.
    \item The obtained pages are distributed in random order to annotators.
    \item Each annotator annotates one by one the pages assigned to them. The annotation of a page proceeds as follows.
    \begin{enumerate}
        \item The annotator writes a question related to the page content, as well as a correct answer to this question (if the answer can be given from the information contained in the page). 
        \item They select one or more spans in the page that allow answering the question.
    \end{enumerate}
    These steps can be repeated to produce up to 5 questions per page.
\end{enumerate}
Annotators were asked to diversify, as much as possible, the type of questions produced, and to indicate the type of each question within a dropdown list.
The possible question types and their numbers are summarized in Table \ref{table:question_types}. Additionally, 15 questions are not associated with any answer, because the necessary information is not available in the business documents. In this case, the system is expected to produce an answer such as "I don't know".

\begin{table}[h]
\centering
\begin{tabularx}{\textwidth}{l c X}
\toprule
\textbf{Question Type} & \textbf{Number} & \textbf{Example} \\
 \midrule
 boolean & 19 & Does a customer of X who cancels due to the November 2022 price increase have to pay fees? \\
 \midrule
 who/what/where/when & 22 & Who should I contact if I have an issue with solution X? \\
 \midrule
 how & 17 & How can I consult the documents and operating procedures for offer X? \\
 \midrule
 why & 14 & Why was the X provided by Y chosen? \\
 \midrule
 conditional & 17 & What happens if a customer wishes to oppose a judicial transfer? \\
 \midrule
 how many & 17 & How many X are included in Y? \\
\bottomrule
\end{tabularx}
\caption{Question types present in the question-answering dataset. Some questions belong to multiple types.}
\label{table:question_types}
\end{table}

The number of spans per question varies from 0 (for questions without answers) to 4, with an average of 1.3.
We observe that the size of spans associated with questions varies greatly, ranging from 1 word (the annotator having selected a single keyword corresponding to the answer) to 202 words. The implications in terms of evaluation metric choice are discussed in Section \ref{sec:metriques}.

\subsection{RAG System Evaluation Procedure}

The question-answering dataset is intended to enable evaluation of the RAG system: first, questions are given as input to the system, and the system outputs (generated responses and retrieved spans) are collected; then the system outputs are evaluated using different metrics.

Comparing generated responses with reference answers produces an overall score, while comparing retrieved spans with reference spans produces a retriever performance score, which can be useful for estimating what proportion of system failures is due to the retriever and which is due to the generator, and thus targeting improvement efforts appropriately.

\subsection{Available Metrics} \label{sec:metriques}

This subsection briefly describes the metrics used to evaluate the RAG system.
We follow the terminology of \citep{ru_ragchecker_2024}, where three types of metrics are distinguished:
\begin{itemize}
    \item \emph{retrieval} metrics, which evaluate the relevance of retrieved spans relative to a given question;
    \item \emph{overall} metrics, which evaluate the response generated by the system for the question;
    \item \emph{generation} metrics, which evaluate the generator's behavior relative to the content of spans from the retrieval.
\end{itemize}

\paragraph{Retrieval Metrics.}
In the context of information retrieval,
several metrics are traditionally used to evaluate retrieval quality: recall, precision, nDCG, MAP, etc.
In this study, we choose to report recall scores, which will be used as reference scores to evaluate retrieval metrics from the tested libraries.
\emph{Recall} is defined as the proportion of relevant elements actually present in the top $k$ spans returned by the retriever. 
A common approach is to calculate recall at the document level, i.e., by considering that a reference span is present in the retriever results if at least one span from the same document is present.
However, this condition does not imply that the reference span is actually contained in a returned span: the reference span can be disjoint from the retrieved span or partially overlapping. This problem arises particularly here, since our references are made of short spans compared to the documents from which they originate. We therefore calculate recall at the word level, i.e.: the percentage of words from reference spans present in the top k retrieved spans. Note that words are identified by their positions in the text, not by their value: returned spans completely disjoint from reference spans will therefore give a score of 0, even if they contain identical words.
The choice of recall over other metrics (such as precision) is justified by the fact that (1) its scores are easy to interpret, (2) since reference spans tend to be relatively small, we seek to know if they are included in the fixed-size spans returned by the retriever, which is a reasonable expectation and corresponds to the definition of recall; conversely, precision at the word level would give necessarily low scores. Finally, the choice of this metric is supported by the fact that it correlates much better with evaluators' average scores ($r=0.35$) than other cited metrics, and notably than document-level recall ($r=0.05$). 

The retrieval metrics proposed by the tested libraries differ from traditional metrics on two main respects: first, some of them do not require reference spans, so they can be applied even when these are not available; second, they often use language models, which allows them to calculate the final score based on semantically important elements of the considered spans. Like recall, these metrics evaluate retrieval quality from the top $k$ retrieved spans. We choose to fix $k=5$ for all metrics, so that the spans on which retrieval is evaluated correspond to those actually inserted into the generator's prompt.

\paragraph{Overall Metrics.} Overall metrics evaluate the quality of the generated response according to specific criteria. Many overall metrics, corresponding to different criteria, are proposed by the tested libraries. Although these criteria are diverse, it can be noted that most are concerned with the factuality of the generated response, generally decomposed into two aspects: \emph{precision} (``is all the information given in the response correct?''), and \emph{recall} (``is all the expected information given in the response?''). In addition to factuality, \emph{relevance} is sometimes considered: ``is all the information given in the response related to the question?'' In addition to metrics evaluating factuality and relevance, we integrate Opik's \emph{moderation} and \emph{usefulness} metrics: the former checks for the absence of harmful or inappropriate content, while the latter combines various criteria to obtain a general score.

\paragraph{Generation Metrics.} Generation metrics evaluate the generator's behavior relative to the content of spans from the retrieval. The primary purpose of these metrics is to study certain generator behaviors and not its performance in the strict sense; however, one may want to use them to approximate overall criteria. For example, DeepEval's \emph{faithfulness} metric, which aims to measure the response's faithfulness to retrieved spans, can be used to approximate the precision criterion. We therefore integrate into our study some generation metrics, which will be evaluated as overall metrics (these are the four metrics named \emph{hallucination} or \emph{faithfulness}, see Figure \ref{fig:1}).

\section{Metric Evaluation: Methodology}\label{sec:method}

We seek to evaluate to what extent the metrics mentioned in the previous section provide a good approximation of a given evaluation criterion. We choose to define an overall criterion simultaneously measuring aspects of factuality and relevance. This criterion is evaluated on a scale of 1 to 5 and defined by the rubric in Table \ref{tab:bareme}.

\begin{table}[h]
\centering
\resizebox{\textwidth}{!}{
\begin{tabularx}{\textwidth}{c X}
\toprule
\textbf{Score} & \textbf{Description} \\
\midrule
5 & Factual and relevant response with the right amount of detail. \\
\midrule
4 & Factual response, but lacking useful information or containing too much unimportant information. \\
\midrule
3 & Partially correct response, with small errors or approximations OR ``I don't know'' when the reference answer contains the requested information. \\
\midrule
2 & Factually incorrect response. \\
\midrule
1 & Off-topic response. \\
\bottomrule
\end{tabularx}
}
\caption{Rubric used during human evaluation of RAG system responses.}
\label{tab:bareme}
\end{table}

This rubric is then applied by two evaluators (company employees with expertise in natural language processing and generative models, one of whom participated in the annotation phase during dataset creation), to score each RAG system output on the 96 instances of the question-answering dataset. We calculate the Pearson correlation\footnote{We calculate Pearson correlation rather than Cohen's or Fleiss's kappa, as these are poorly suited when annotations are ordinal in nature, as is the case here.} between the two series of scores obtained, to verify the robustness of the rubric and estimate the maximum performance that can be expected from an automatic metric. The correlation obtained is 0.85.
To simplify subsequent analyses, we base ourselves on the average scores obtained per response. The scores thus obtained are called \emph{reference scores}.

We then proceed to the correlation analysis intended to evaluate the metrics. The following section reports and analyzes:
\begin{itemize}
    \item correlations between overall metrics and reference scores;
    \item correlations of retrieval metrics with recall: although recall is itself an imperfect metric, it is expected that a stronger correlation of a retrieval metric with recall indicates better reliability;
    \item correlations between retrieval metrics and reference scores: indeed, since better retrieval correlates with better responses, retrieval metrics should also correlate with scores from our rubric.
\end{itemize}
All reported correlations correspond to Pearson's coefficient (performing analyses with Spearman correlation gives results similar to those reported).

\section{Results}\label{sec:results}

\input{fig1}

Figure \ref{fig:1} summarizes the correlations obtained for overall metrics. We first note that the width of confidence intervals is substantial, due to the modest size of our sample (96). However, it is possible to see certain trends emerge.

First, we note that metrics sometimes considered obsolete such as METEOR correlate surprisingly well with reference scores. Next, we observe that generation metrics used as overall metrics correlate weakly with reference scores, which is expected, since they do not have access to the reference answer. Similarly, Opik's \emph{moderation} metric does not correlate with reference scores.
Conversely, we observe a very strong correlation for RAGChecker metrics, particularly recall.

\input{fig2}

Figure \ref{fig:2} summarizes the correlations of retrieval metrics with recall. We observe larger overlaps of confidence intervals. However, it is interesting to note that a metric such as DeepEval's \emph{contextual precision} obtains a correlation above 0.5 with recall, without using reference spans. We observe that conversely, Ragas's non-LLM metrics obtain very weak correlation. This is probably explained by the fact that it compares retrieved spans to reference spans via Levenshtein distance, which does not give a relevant value when compared spans are of very different sizes, as is the case here.

\input{fig3}

If we observe Figure \ref{fig:3}, we note a surprising phenomenon: several metrics show very strong correlation with reference scores. This correlation is sometimes stronger than the correlation with recall. RAGChecker's \emph{claim recall} correlation appears close to 0.7. It does not seem plausible that such a correlation is due solely to this metric's ability to measure the relevance of retrieved documents. We comment on this phenomenon in more detail in the following section.

\section{Limitations}\label{sec:limites}

The main limitation of our study is related to the interpretation of correlations. It is difficult to know a priori what sophisticated metrics using LLMs rely on to produce their scores. Consequently, observing that a given metric correlates well with human judgment is not sufficient to affirm that it measures what we want it to measure, given the configuration of our experiment. This difficulty can be illustrated by imagining a metric measuring question difficulty, without knowledge of generated responses. Such a metric would give higher scores to easy questions, and these indeed tend to obtain better scores. It would therefore have a positive correlation with reference scores, while its relevance for evaluating and comparing RAG systems would be null (it would give exactly the same scores to all systems). A less extreme illustration of this phenomenon is probably RAGChecker's \emph{claim recall} metric, whose correlation with evaluators' scores is very high. It seems likely that this metric does not measure only retrieval quality; a plausible interpretation is that it partially captures other characteristics such as, for example, the ease with which response information can be extracted from retrieved spans, or question difficulty. This limitation is partly due to the configuration of our experiment: since it involves only one RAG system, it is impossible to observe metric behavior as a function of system outputs independently of the input question.

Note that, despite this weakness, our methodology produces certain exploitable results by eliminating certain candidate metrics: indeed, we can affirm that metrics with poor correlation with the evaluation criterion considered do not measure this criterion.

\section{Related Work}\label{sec:sota}
Metric evaluations through correlation studies with human judgment have been reported in several recent publications. Although most of them consider the evaluation of NLG (Natural Language Generation) tasks other than RAG, the evaluation of these tasks involves issues common to those of evaluating RAG system responses. This section offers a (non-exhaustive) overview of these publications, grouped according to their evaluation methodologies.

First, some of these publications rely on a methodology similar to ours (correlation analysis on a single system). We can cite, for example: \citep{liu_x-eval_2024} which evaluates a system dedicated to evaluating various aspects of various natural language generation tasks, or \citep{yeginbergen_dynamic_2025} which observes the correlation of several LLM-as-a-judge systems with human evaluations of automatically generated counter-arguments. These studies suffer from the same methodological limitation as ours.

Other studies evaluate metrics by measuring their adequacy with a preference order expressed on pairs of outputs corresponding to the same input. More precisely: each task input is associated with two alternative outputs, for which a preference order is available; all pairs for which the metric score conforms to the preference order are considered successes, and the success ratio forms the metric's performance score. This approach helps guard against confounding factors related to question characteristics that limit the interpretations of our results. Among studies using this approach, we can cite: \citep{ke_critiquellm_2024,zhu_judgelm_2025,lambert_rewardbench_2025}. Some studies combine this approach with correlation analysis on a single system, for example: \citep{xu_instructscore_2023,kim_prometheus_2024, xiong_lllava-critic_2025}.

Finally, some publications report correlation analyses involving multiple systems. The statistical treatments performed can then vary, since the scores generated by a metric, like reference scores, are then arranged into a matrix with one row per system and one column per evaluated output. There are indeed several ways to calculate a correlation between two matrices: \citep{gao_analyzing_2025} studies four of them, two of which seem relevant to us in the context of evaluating RAG metrics. The first is to calculate the correlation of each system's average scores noted by the metric with each system's average reference score. The second is to calculate, for each input, the correlation of metric scores with reference scores across different systems, then to average the correlations thus obtained. These two approaches are also studied by \citep{deutsch_statistical_2021}. Both studies empirically conclude that the second approach has greater power to discriminate among tested metrics. Moreover, it has been demonstrated that this approach considerably reduces the importance of confounding factors in correlations measured between machine translation evaluation metrics and human judgment \citep{perrella_guardians_2024}. A similar approach, applied in \citep{dinh_sciex_2024}, consists of calculating correlation on all scores, normalized by inputs.

\section{Research Directions}\label{sec:pistes}

The empirical results of our study are in line with some of the results mentioned in the previous section, and confirm the usefulness of integrating several (at least two) RAG systems into correlation studies aimed at evaluating RAG metrics. It should be noted that the study results will then be partially dependent on the chosen RAG systems: it is therefore appropriate to choose a set of systems representative of all systems to which we wish to apply the metrics. Moreover, integrating too many systems risks making the annotation task costly.

These difficulties open interesting research perspectives. For example, how to estimate the probable contribution of an additional model or question to a correlation study? Is it possible to develop procedures to choose and adapt the number of systems and questions to integrate into the empirical study, as human evaluations are collected?
Answers to such questions could enable professionals implementing RAG systems to validate their evaluation metrics more reliably while minimizing the costs associated with this validation.

\bibliographystyle{plainnat}
\bibliography{biblio}

\end{document}

%% file: fig1.tex
\begin{figure}[h]
        \centering    \begin{tikzpicture}
      \draw[gray!30, solid, thin] (4.351, 0) -- (4.351, 10.800);
      \draw[gray!30, solid, thin] (5.406, 0) -- (5.406, 10.800);
      \draw[gray!30, solid, thin] (6.460, 0) -- (6.460, 10.800);
      \draw[gray!30, solid, thin] (7.515, 0) -- (7.515, 10.800);
      \draw[gray!30, dashed, thin] (8.570, 0) -- (8.570, 10.800);
      \draw[gray!30, solid, thin] (9.625, 0) -- (9.625, 10.800);
      \draw[gray!30, solid, thin] (10.680, 0) -- (10.680, 10.800);
      \draw[gray!30, solid, thin] (11.734, 0) -- (11.734, 10.800);
      \draw[gray!30, solid, thin] (12.789, 0) -- (12.789, 10.800);
      \draw[black, dashed, thin] (8.570, 0) -- (8.570, 10.800);
      \draw[black] (4.140, 0) -- (13.000, 0);
      \draw (4.351, 0) -- (4.351, -0.15) node[below, font=\small] {-1.00};
      \draw (5.406, 0) -- (5.406, -0.15) node[below, font=\small] {-0.75};
      \draw (6.460, 0) -- (6.460, -0.15) node[below, font=\small] {-0.50};
      \draw (7.515, 0) -- (7.515, -0.15) node[below, font=\small] {-0.25};
      \draw (8.570, 0) -- (8.570, -0.15) node[below, font=\small] {0.00};
      \draw (9.625, 0) -- (9.625, -0.15) node[below, font=\small] {0.25};
      \draw (10.680, 0) -- (10.680, -0.15) node[below, font=\small] {0.50};
      \draw (11.734, 0) -- (11.734, -0.15) node[below, font=\small] {0.75};
      \draw (12.789, 0) -- (12.789, -0.15) node[below, font=\small] {1.00};
      \draw[gray!40, dashed, thin] (-0.6, 8.700) -- (13.000, 8.700);
      \node[rotate=90, font=\small\bfseries, text width=1.500cm, align=center] at (-0.300, 7.850) {DeepEval};
      \draw[gray!60, thin] (0.000, 7.100) -- (0.000, 8.600);
      \draw[gray!40, dashed, thin] (-0.6, 7.000) -- (13.000, 7.000);
      \node[rotate=90, font=\small\bfseries, text width=2.000cm, align=center] at (-0.300, 5.900) {Ragas};
      \draw[gray!60, thin] (0.000, 4.900) -- (0.000, 6.900);
      \draw[gray!40, dashed, thin] (-0.6, 4.800) -- (13.000, 4.800);
      \node[rotate=90, font=\small\bfseries, text width=3.000cm, align=center] at (-0.300, 3.200) {Opik};
      \draw[gray!60, thin] (0.000, 1.700) -- (0.000, 4.700);
      \draw[gray!40, dashed, thin] (-0.6, 1.600) -- (13.000, 1.600);
      \node[rotate=90, font=\small\bfseries, text width=1.500cm, align=center] at (-0.400, 0.750) {RAG\\Checker};
      \draw[gray!60, thin] (0.000, 0.000) -- (0.000, 1.500);
      \node[left, font=\footnotesize] at (4.140, 10.550) {bleu};
      \draw[black, thick] (9.108, 10.550) -- (10.647, 10.550);
      \draw[black, thick] (9.108, 10.400) -- (9.108, 10.700);
      \draw[black, thick] (10.647, 10.400) -- (10.647, 10.700);
      \filldraw[black] (9.928, 10.550) circle (0.06cm);
      \node[left, font=\footnotesize] at (4.140, 10.050) {rougeLsum};
      \draw[black, thick] (9.837, 10.050) -- (11.175, 10.050);
      \draw[black, thick] (9.837, 9.900) -- (9.837, 10.200);
      \draw[black, thick] (11.175, 9.900) -- (11.175, 10.200);
      \filldraw[black] (10.570, 10.050) circle (0.06cm);
      \node[left, font=\footnotesize] at (4.140, 9.550) {meteor};
      \draw[black, thick] (10.251, 9.550) -- (11.447, 9.550);
      \draw[black, thick] (10.251, 9.400) -- (10.251, 9.700);
      \draw[black, thick] (11.447, 9.400) -- (11.447, 9.700);
      \filldraw[black] (10.916, 9.550) circle (0.06cm);
      \node[left, font=\footnotesize] at (4.140, 9.050) {bertscore-f1};
      \draw[black, thick] (10.061, 9.050) -- (11.324, 9.050);
      \draw[black, thick] (10.061, 8.900) -- (10.061, 9.200);
      \draw[black, thick] (11.324, 8.900) -- (11.324, 9.200);
      \filldraw[black] (10.759, 9.050) circle (0.06cm);
      \node[left, font=\footnotesize] at (4.140, 8.350) {hallucination};
      \draw[black, thick] (6.531, 8.350) -- (8.082, 8.350);
      \draw[black, thick] (6.531, 8.200) -- (6.531, 8.500);
      \draw[black, thick] (8.082, 8.200) -- (8.082, 8.500);
      \filldraw[black] (7.258, 8.350) circle (0.06cm);
      \node[left, font=\footnotesize] at (4.140, 7.850) {answer relevancy};
      \draw[black, thick] (9.260, 7.850) -- (10.763, 7.850);
      \draw[black, thick] (9.260, 7.700) -- (9.260, 8.000);
      \draw[black, thick] (10.763, 7.700) -- (10.763, 8.000);
      \filldraw[black] (10.066, 7.850) circle (0.06cm);
      \node[left, font=\footnotesize] at (4.140, 7.350) {faithfulness};
      \draw[black, thick] (7.927, 7.350) -- (9.632, 7.350);
      \draw[black, thick] (7.927, 7.200) -- (7.927, 7.500);
      \draw[black, thick] (9.632, 7.200) -- (9.632, 7.500);
      \filldraw[black] (8.788, 7.350) circle (0.06cm);
      \node[left, font=\footnotesize] at (4.140, 6.650) {faithfulness};
      \draw[black, thick] (8.269, 6.650) -- (9.950, 6.650);
      \draw[black, thick] (8.269, 6.500) -- (8.269, 6.800);
      \draw[black, thick] (9.950, 6.500) -- (9.950, 6.800);
      \filldraw[black] (9.132, 6.650) circle (0.06cm);
      \node[left, font=\footnotesize] at (4.140, 6.150) {factual correctness:f1};
      \draw[black, thick] (9.384, 6.150) -- (10.855, 6.150);
      \draw[black, thick] (9.384, 6.000) -- (9.384, 6.300);
      \draw[black, thick] (10.855, 6.000) -- (10.855, 6.300);
      \filldraw[black] (10.176, 6.150) circle (0.06cm);
      \node[left, font=\footnotesize] at (4.140, 5.650) {factual correctness:recall};
      \draw[black, thick] (9.444, 5.650) -- (10.899, 5.650);
      \draw[black, thick] (9.444, 5.500) -- (9.444, 5.800);
      \draw[black, thick] (10.899, 5.500) -- (10.899, 5.800);
      \filldraw[black] (10.229, 5.650) circle (0.06cm);
      \node[left, font=\footnotesize] at (4.140, 5.150) {factual correctness:accuracy};
      \draw[black, thick] (9.384, 5.150) -- (10.855, 5.150);
      \draw[black, thick] (9.384, 5.000) -- (9.384, 5.300);
      \draw[black, thick] (10.855, 5.000) -- (10.855, 5.300);
      \filldraw[black] (10.176, 5.150) circle (0.06cm);
      \node[left, font=\footnotesize] at (4.140, 4.450) {hallucination};
      \draw[black, thick] (7.130, 4.450) -- (8.803, 4.450);
      \draw[black, thick] (7.130, 4.300) -- (7.130, 4.600);
      \draw[black, thick] (8.803, 4.300) -- (8.803, 4.600);
      \filldraw[black] (7.941, 4.450) circle (0.06cm);
      \node[left, font=\footnotesize] at (4.140, 3.950) {moderation};
      \draw[black, thick] (7.526, 3.950) -- (9.232, 3.950);
      \draw[black, thick] (7.526, 3.800) -- (7.526, 4.100);
      \draw[black, thick] (9.232, 3.800) -- (9.232, 4.100);
      \filldraw[black] (8.371, 3.950) circle (0.06cm);
      \node[left, font=\footnotesize] at (4.140, 3.450) {answer relevance};
      \draw[black, thick] (9.189, 3.450) -- (10.709, 3.450);
      \draw[black, thick] (9.189, 3.300) -- (9.189, 3.600);
      \draw[black, thick] (10.709, 3.300) -- (10.709, 3.600);
      \filldraw[black] (10.001, 3.450) circle (0.06cm);
      \node[left, font=\footnotesize] at (4.140, 2.950) {usefulness};
      \draw[black, thick] (8.493, 2.950) -- (10.146, 2.950);
      \draw[black, thick] (8.493, 2.800) -- (8.493, 3.100);
      \draw[black, thick] (10.146, 2.800) -- (10.146, 3.100);
      \filldraw[black] (9.351, 2.950) circle (0.06cm);
      \node[left, font=\footnotesize] at (4.140, 2.450) {context precision};
      \draw[black, thick] (10.136, 2.450) -- (11.373, 2.450);
      \draw[black, thick] (10.136, 2.300) -- (10.136, 2.600);
      \draw[black, thick] (11.373, 2.300) -- (11.373, 2.600);
      \filldraw[black] (10.821, 2.450) circle (0.06cm);
      \node[left, font=\footnotesize] at (4.140, 1.950) {context recall};
      \draw[black, thick] (10.030, 1.950) -- (11.304, 1.950);
      \draw[black, thick] (10.030, 1.800) -- (10.030, 2.100);
      \draw[black, thick] (11.304, 1.800) -- (11.304, 2.100);
      \filldraw[black] (10.733, 1.950) circle (0.06cm);
      \node[left, font=\footnotesize] at (4.140, 1.250) {f1};
      \draw[black, thick] (11.156, 1.250) -- (11.984, 1.250);
      \draw[black, thick] (11.156, 1.100) -- (11.156, 1.400);
      \draw[black, thick] (11.984, 1.100) -- (11.984, 1.400);
      \filldraw[black] (11.631, 1.250) circle (0.06cm);
      \node[left, font=\footnotesize] at (4.140, 0.750) {recall};
      \draw[black, thick] (11.484, 0.750) -- (12.161, 0.750);
      \draw[black, thick] (11.484, 0.600) -- (11.484, 0.900);
      \draw[black, thick] (12.161, 0.600) -- (12.161, 0.900);
      \filldraw[black] (11.876, 0.750) circle (0.06cm);
      \node[left, font=\footnotesize] at (4.140, 0.250) {precision};
      \draw[black, thick] (10.509, 0.250) -- (11.608, 0.250);
      \draw[black, thick] (10.509, 0.100) -- (10.509, 0.400);
      \draw[black, thick] (11.608, 0.100) -- (11.608, 0.400);
      \filldraw[black] (11.126, 0.250) circle (0.06cm);
    \end{tikzpicture}\caption{Pearson correlation of overall metrics with average human rates.}
    \label{fig:1}
    \end{figure}
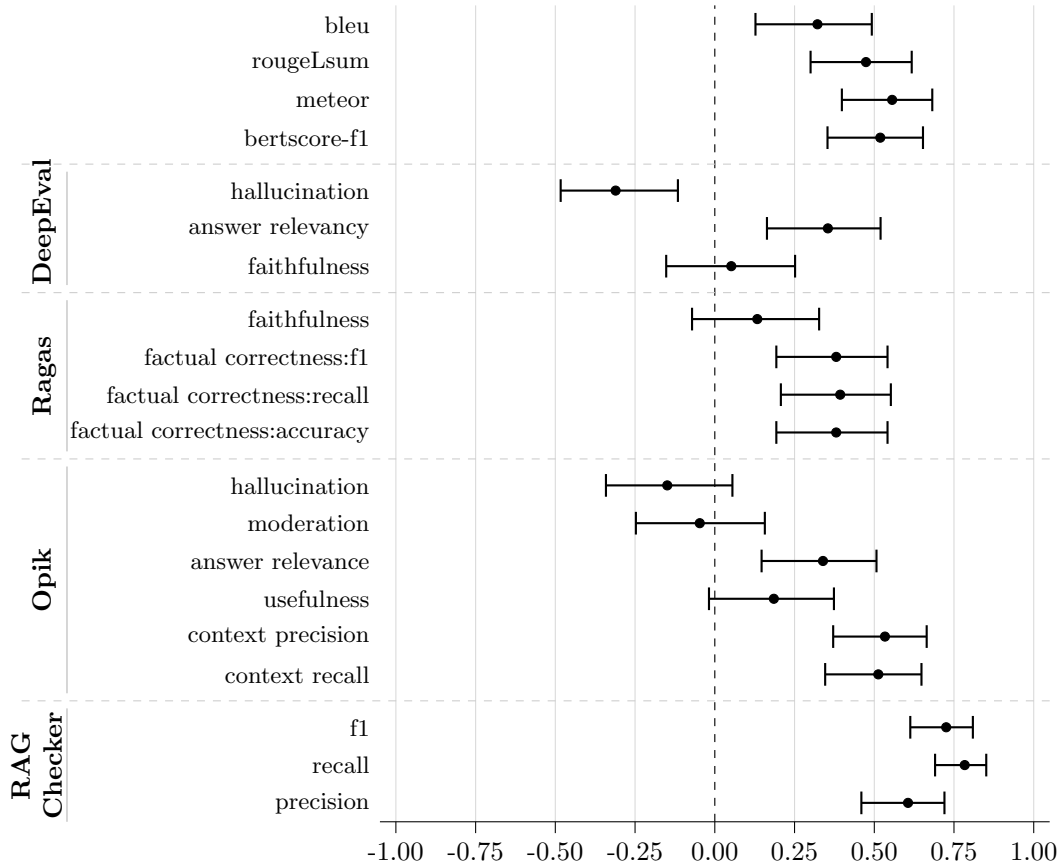

%% file: fig2.tex
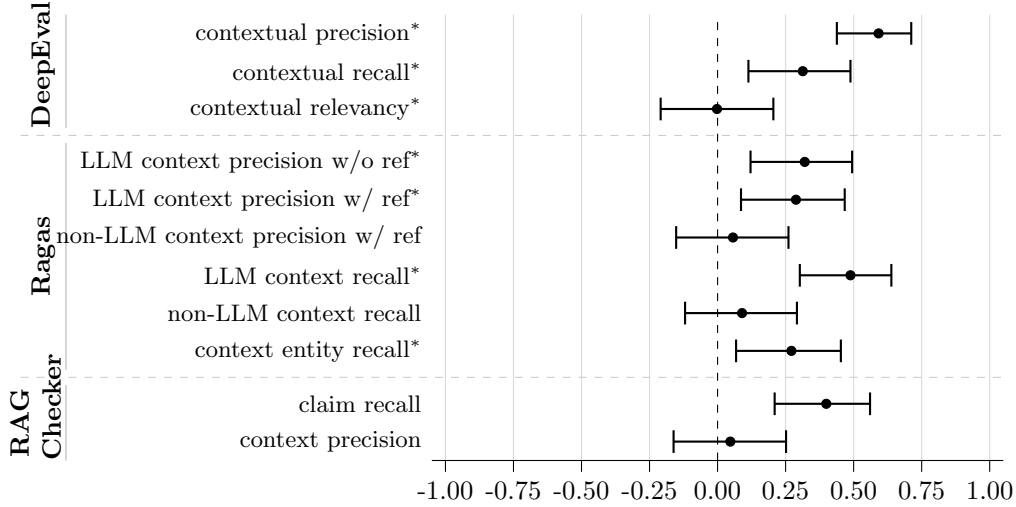
\begin{figure}[h]
        \centering    \begin{tikzpicture}
      \draw[gray!30, solid, thin] (5.620, 0) -- (5.620, 5.900);
      \draw[gray!30, solid, thin] (6.520, 0) -- (6.520, 5.900);
      \draw[gray!30, solid, thin] (7.420, 0) -- (7.420, 5.900);
      \draw[gray!30, solid, thin] (8.320, 0) -- (8.320, 5.900);
      \draw[gray!30, dashed, thin] (9.220, 0) -- (9.220, 5.900);
      \draw[gray!30, solid, thin] (10.120, 0) -- (10.120, 5.900);
      \draw[gray!30, solid, thin] (11.020, 0) -- (11.020, 5.900);
      \draw[gray!30, solid, thin] (11.920, 0) -- (11.920, 5.900);
      \draw[gray!30, solid, thin] (12.820, 0) -- (12.820, 5.900);
      \draw[black, dashed, thin] (9.220, 0) -- (9.220, 5.900);
      \draw[black] (5.440, 0) -- (13.000, 0);
      \draw (5.620, 0) -- (5.620, -0.15) node[below, font=\small] {-1.00};
      \draw (6.520, 0) -- (6.520, -0.15) node[below, font=\small] {-0.75};
      \draw (7.420, 0) -- (7.420, -0.15) node[below, font=\small] {-0.50};
      \draw (8.320, 0) -- (8.320, -0.15) node[below, font=\small] {-0.25};
      \draw (9.220, 0) -- (9.220, -0.15) node[below, font=\small] {0.00};
      \draw (10.120, 0) -- (10.120, -0.15) node[below, font=\small] {0.25};
      \draw (11.020, 0) -- (11.020, -0.15) node[below, font=\small] {0.50};
      \draw (11.920, 0) -- (11.920, -0.15) node[below, font=\small] {0.75};
      \draw (12.820, 0) -- (12.820, -0.15) node[below, font=\small] {1.00};
      \node[rotate=90, font=\small\bfseries, text width=1.500cm, align=center] at (0.300, 5.150) {DeepEval};
      \draw[gray!60, thin] (0.600, 4.400) -- (0.600, 5.900);
      \draw[gray!40, dashed, thin] (0, 4.300) -- (13.000, 4.300);
      \node[rotate=90, font=\small\bfseries, text width=3.000cm, align=center] at (0.300, 2.700) {Ragas};
      \draw[gray!60, thin] (0.600, 1.200) -- (0.600, 4.200);
      \draw[gray!40, dashed, thin] (0, 1.100) -- (13.000, 1.100);
      \node[rotate=90, font=\small\bfseries, text width=1.000cm, align=center] at (0.200, 0.500) {RAG\\Checker};
      \draw[gray!60, thin] (0.600, 0.000) -- (0.600, 1.000);
      \node[left, font=\footnotesize] at (5.440, 5.650) {contextual precision$^*$};
      \draw[black, thick] (10.798, 5.650) -- (11.782, 5.650);
      \draw[black, thick] (10.798, 5.500) -- (10.798, 5.800);
      \draw[black, thick] (11.782, 5.500) -- (11.782, 5.800);
      \filldraw[black] (11.350, 5.650) circle (0.06cm);
      \node[left, font=\footnotesize] at (5.440, 5.150) {contextual recall$^*$};
      \draw[black, thick] (9.628, 5.150) -- (10.978, 5.150);
      \draw[black, thick] (9.628, 5.000) -- (9.628, 5.300);
      \draw[black, thick] (10.978, 5.000) -- (10.978, 5.300);
      \filldraw[black] (10.347, 5.150) circle (0.06cm);
      \node[left, font=\footnotesize] at (5.440, 4.650) {contextual relevancy$^*$};
      \draw[black, thick] (8.467, 4.650) -- (9.958, 4.650);
      \draw[black, thick] (8.467, 4.500) -- (8.467, 4.800);
      \draw[black, thick] (9.958, 4.500) -- (9.958, 4.800);
      \filldraw[black] (9.212, 4.650) circle (0.06cm);
      \node[left, font=\footnotesize] at (5.440, 3.950) {LLM context precision w/o ref$^*$};
      \draw[black, thick] (9.656, 3.950) -- (11.000, 3.950);
      \draw[black, thick] (9.656, 3.800) -- (9.656, 4.100);
      \draw[black, thick] (11.000, 3.800) -- (11.000, 4.100);
      \filldraw[black] (10.373, 3.950) circle (0.06cm);
      \node[left, font=\footnotesize] at (5.440, 3.450) {LLM context precision w/ ref$^*$};
      \draw[black, thick] (9.530, 3.450) -- (10.902, 3.450);
      \draw[black, thick] (9.530, 3.300) -- (9.530, 3.600);
      \draw[black, thick] (10.902, 3.300) -- (10.902, 3.600);
      \filldraw[black] (10.257, 3.450) circle (0.06cm);
      \node[left, font=\footnotesize] at (5.440, 2.950) {non-LLM context precision w/ ref};
      \draw[black, thick] (8.672, 2.950) -- (10.158, 2.950);
      \draw[black, thick] (8.672, 2.800) -- (8.672, 3.100);
      \draw[black, thick] (10.158, 2.800) -- (10.158, 3.100);
      \filldraw[black] (9.424, 2.950) circle (0.06cm);
      \node[left, font=\footnotesize] at (5.440, 2.450) {LLM context recall$^*$};
      \draw[black, thick] (10.308, 2.450) -- (11.519, 2.450);
      \draw[black, thick] (10.308, 2.300) -- (10.308, 2.600);
      \draw[black, thick] (11.519, 2.300) -- (11.519, 2.600);
      \filldraw[black] (10.978, 2.450) circle (0.06cm);
      \node[left, font=\footnotesize] at (5.440, 1.950) {non-LLM context recall};
      \draw[black, thick] (8.790, 1.950) -- (10.270, 1.950);
      \draw[black, thick] (8.790, 1.800) -- (8.790, 2.100);
      \draw[black, thick] (10.270, 1.800) -- (10.270, 2.100);
      \filldraw[black] (9.544, 1.950) circle (0.06cm);
      \node[left, font=\footnotesize] at (5.440, 1.450) {context entity recall$^*$};
      \draw[black, thick] (9.465, 1.450) -- (10.851, 1.450);
      \draw[black, thick] (9.465, 1.300) -- (9.465, 1.600);
      \draw[black, thick] (10.851, 1.300) -- (10.851, 1.600);
      \filldraw[black] (10.197, 1.450) circle (0.06cm);
      \node[left, font=\footnotesize] at (5.440, 0.750) {claim recall};
      \draw[black, thick] (9.975, 0.750) -- (11.237, 0.750);
      \draw[black, thick] (9.975, 0.600) -- (9.975, 0.900);
      \draw[black, thick] (11.237, 0.600) -- (11.237, 0.900);
      \filldraw[black] (10.658, 0.750) circle (0.06cm);
      \node[left, font=\footnotesize] at (5.440, 0.250) {context precision};
      \draw[black, thick] (8.638, 0.250) -- (10.126, 0.250);
      \draw[black, thick] (8.638, 0.100) -- (8.638, 0.400);
      \draw[black, thick] (10.126, 0.100) -- (10.126, 0.400);
      \filldraw[black] (9.389, 0.250) circle (0.06cm);
    \end{tikzpicture}\caption{Pearson correlation of overall metrics with recall. Metrics that do not rely of reference spans are marked with an asterisk.}
    \label{fig:2}
    \end{figure}

%% file: fig3.tex
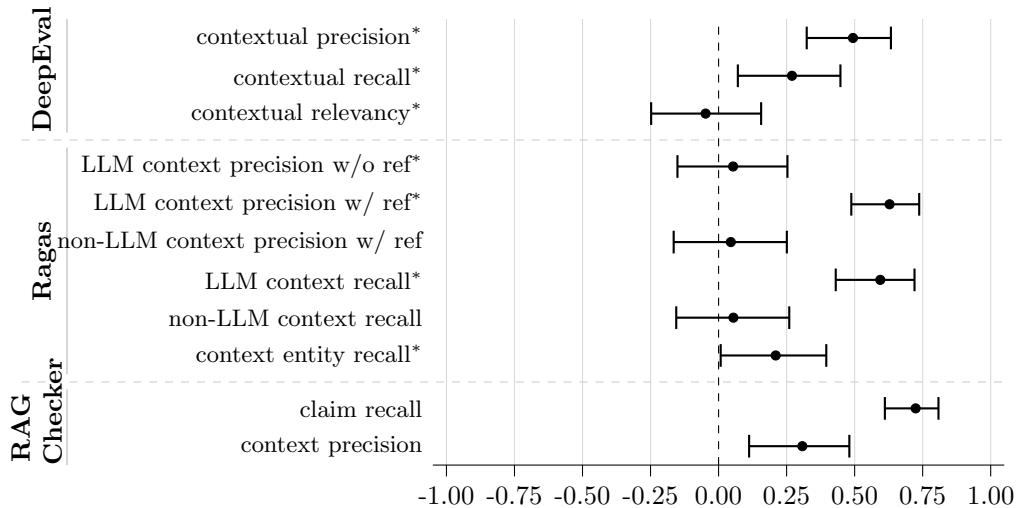
\begin{figure}[h]
        \centering    \begin{tikzpicture}
      \draw[gray!30, solid, thin] (5.620, 0) -- (5.620, 5.900);
      \draw[gray!30, solid, thin] (6.520, 0) -- (6.520, 5.900);
      \draw[gray!30, solid, thin] (7.420, 0) -- (7.420, 5.900);
      \draw[gray!30, solid, thin] (8.320, 0) -- (8.320, 5.900);
      \draw[gray!30, dashed, thin] (9.220, 0) -- (9.220, 5.900);
      \draw[gray!30, solid, thin] (10.120, 0) -- (10.120, 5.900);
      \draw[gray!30, solid, thin] (11.020, 0) -- (11.020, 5.900);
      \draw[gray!30, solid, thin] (11.920, 0) -- (11.920, 5.900);
      \draw[gray!30, solid, thin] (12.820, 0) -- (12.820, 5.900);
      \draw[black, dashed, thin] (9.220, 0) -- (9.220, 5.900);
      \draw[black] (5.440, 0) -- (13.000, 0);
      \draw (5.620, 0) -- (5.620, -0.15) node[below, font=\small] {-1.00};
      \draw (6.520, 0) -- (6.520, -0.15) node[below, font=\small] {-0.75};
      \draw (7.420, 0) -- (7.420, -0.15) node[below, font=\small] {-0.50};
      \draw (8.320, 0) -- (8.320, -0.15) node[below, font=\small] {-0.25};
      \draw (9.220, 0) -- (9.220, -0.15) node[below, font=\small] {0.00};
      \draw (10.120, 0) -- (10.120, -0.15) node[below, font=\small] {0.25};
      \draw (11.020, 0) -- (11.020, -0.15) node[below, font=\small] {0.50};
      \draw (11.920, 0) -- (11.920, -0.15) node[below, font=\small] {0.75};
      \draw (12.820, 0) -- (12.820, -0.15) node[below, font=\small] {1.00};
      \node[rotate=90, font=\small\bfseries, text width=1.500cm, align=center] at (0.300, 5.150) {DeepEval};
      \draw[gray!60, thin] (0.600, 4.400) -- (0.600, 5.900);
      \draw[gray!40, dashed, thin] (0, 4.300) -- (13.000, 4.300);
      \node[rotate=90, font=\small\bfseries, text width=3.000cm, align=center] at (0.300, 2.700) {Ragas};
      \draw[gray!60, thin] (0.600, 1.200) -- (0.600, 4.200);
      \draw[gray!40, dashed, thin] (0, 1.100) -- (13.000, 1.100);
      \node[rotate=90, font=\small\bfseries, text width=1.000cm, align=center] at (0.200, 0.500) {RAG\\Checker};
      \draw[gray!60, thin] (0.600, 0.000) -- (0.600, 1.000);
      \node[left, font=\footnotesize] at (5.440, 5.650) {contextual precision$^*$};
      \draw[black, thick] (10.386, 5.650) -- (11.500, 5.650);
      \draw[black, thick] (10.386, 5.500) -- (10.386, 5.800);
      \draw[black, thick] (11.500, 5.500) -- (11.500, 5.800);
      \filldraw[black] (10.998, 5.650) circle (0.06cm);
      \node[left, font=\footnotesize] at (5.440, 5.150) {contextual recall$^*$};
      \draw[black, thick] (9.475, 5.150) -- (10.832, 5.150);
      \draw[black, thick] (9.475, 5.000) -- (9.475, 5.300);
      \draw[black, thick] (10.832, 5.000) -- (10.832, 5.300);
      \filldraw[black] (10.191, 5.150) circle (0.06cm);
      \node[left, font=\footnotesize] at (5.440, 4.650) {contextual relevancy$^*$};
      \draw[black, thick] (8.328, 4.650) -- (9.783, 4.650);
      \draw[black, thick] (8.328, 4.500) -- (8.328, 4.800);
      \draw[black, thick] (9.783, 4.500) -- (9.783, 4.800);
      \filldraw[black] (9.048, 4.650) circle (0.06cm);
      \node[left, font=\footnotesize] at (5.440, 3.950) {LLM context precision w/o ref$^*$};
      \draw[black, thick] (8.676, 3.950) -- (10.131, 3.950);
      \draw[black, thick] (8.676, 3.800) -- (8.676, 4.100);
      \draw[black, thick] (10.131, 3.800) -- (10.131, 4.100);
      \filldraw[black] (9.412, 3.950) circle (0.06cm);
      \node[left, font=\footnotesize] at (5.440, 3.450) {LLM context precision w/ ref$^*$};
      \draw[black, thick] (10.976, 3.450) -- (11.874, 3.450);
      \draw[black, thick] (10.976, 3.300) -- (10.976, 3.600);
      \draw[black, thick] (11.874, 3.300) -- (11.874, 3.600);
      \filldraw[black] (11.482, 3.450) circle (0.06cm);
      \node[left, font=\footnotesize] at (5.440, 2.950) {non-LLM context precision w/ ref};
      \draw[black, thick] (8.626, 2.950) -- (10.123, 2.950);
      \draw[black, thick] (8.626, 2.800) -- (8.626, 3.100);
      \draw[black, thick] (10.123, 2.800) -- (10.123, 3.100);
      \filldraw[black] (9.382, 2.950) circle (0.06cm);
      \node[left, font=\footnotesize] at (5.440, 2.450) {LLM context recall$^*$};
      \draw[black, thick] (10.770, 2.450) -- (11.812, 2.450);
      \draw[black, thick] (10.770, 2.300) -- (10.770, 2.600);
      \draw[black, thick] (11.812, 2.300) -- (11.812, 2.600);
      \filldraw[black] (11.359, 2.450) circle (0.06cm);
      \node[left, font=\footnotesize] at (5.440, 1.950) {non-LLM context recall};
      \draw[black, thick] (8.660, 1.950) -- (10.155, 1.950);
      \draw[black, thick] (8.660, 1.800) -- (8.660, 2.100);
      \draw[black, thick] (10.155, 1.800) -- (10.155, 2.100);
      \filldraw[black] (9.416, 1.950) circle (0.06cm);
      \node[left, font=\footnotesize] at (5.440, 1.450) {context entity recall$^*$};
      \draw[black, thick] (9.248, 1.450) -- (10.645, 1.450);
      \draw[black, thick] (9.248, 1.300) -- (9.248, 1.600);
      \draw[black, thick] (10.645, 1.300) -- (10.645, 1.600);
      \filldraw[black] (9.976, 1.450) circle (0.06cm);
      \node[left, font=\footnotesize] at (5.440, 0.750) {claim recall};
      \draw[black, thick] (11.420, 0.750) -- (12.129, 0.750);
      \draw[black, thick] (11.420, 0.600) -- (11.420, 0.900);
      \draw[black, thick] (12.129, 0.600) -- (12.129, 0.900);
      \filldraw[black] (11.827, 0.750) circle (0.06cm);
      \node[left, font=\footnotesize] at (5.440, 0.250) {context precision};
      \draw[black, thick] (9.624, 0.250) -- (10.950, 0.250);
      \draw[black, thick] (9.624, 0.100) -- (9.624, 0.400);
      \draw[black, thick] (10.950, 0.100) -- (10.950, 0.400);
      \filldraw[black] (10.328, 0.250) circle (0.06cm);
    \end{tikzpicture}\caption{Pearson correlations of retrieval metrics with average human rates. Metrics that do not rely on reference spans are marked with an asterisk.}
    \label{fig:3}
    \end{figure}

%% file: biblio.bib
@inproceedings{ru_ragchecker_2024,
	title = {{RAGChecker}: {A} {Fine}-grained {Framework} for {Diagnosing} {Retrieval}-{Augmented} {Generation}},
	shorttitle = {{RAGChecker}},
	url = {http://arxiv.org/abs/2408.08067},
	doi = {10.48550/arXiv.2408.08067},
	author = {Ru, Dongyu and Qiu, Lin and Hu, Xiangkun and Zhang, Tianhang and Shi, Peng and Chang, Shuaichen and Jiayang, Cheng and Wang, Cunxiang and Sun, Shichao and Li, Huanyu and Zhang, Zizhao and Wang, Binjie and Jiang, Jiarong and He, Tong and Wang, Zhiguo and Liu, Pengfei and Zhang, Yue and Zhang, Zheng},
	year = {2024}
}

@inproceedings{es_ragas_2024,
	address = {St. Julians, Malta},
	title = {{RAGAs}: {Automated} {Evaluation} of {Retrieval} {Augmented} {Generation}},
	shorttitle = {{RAGAs}},
	url = {https://aclanthology.org/2024.eacl-demo.16/},
	doi = {10.18653/v1/2024.eacl-demo.16},
	booktitle = {Proceedings of the 18th {Conference} of the {European} {Chapter} of the {Association} for {Computational} {Linguistics}: {System} {Demonstrations}},
	publisher = {Association for Computational Linguistics},
	author = {Es, Shahul and James, Jithin and Espinosa Anke, Luis and Schockaert, Steven},
	editor = {Aletras, Nikolaos and De Clercq, Orphee},
	month = mar,
	year = {2024},
	pages = {150--158}
}

@article{deutsch_statistical_2021,
	title = {A {Statistical} {Analysis} of {Summarization} {Evaluation} {Metrics} {Using} {Resampling} {Methods}},
	volume = {9},
	issn = {2307-387X},
	url = {https://doi.org/10.1162/tacl_a_00417},
	doi = {10.1162/tacl_a_00417},
	journal = {Transactions of the Association for Computational Linguistics},
	author = {Deutsch, Daniel and Dror, Rotem and Roth, Dan},
	month = oct,
	year = {2021},
	pages = {1132--1146}
}

@inproceedings{gao_analyzing_2025,
	address = {Albuquerque, New Mexico},
	title = {Analyzing and {Evaluating} {Correlation} {Measures} in {NLG} {Meta}-{Evaluation}},
	isbn = {979-8-89176-189-6},
	url = {https://aclanthology.org/2025.naacl-long.111/},
	doi = {10.18653/v1/2025.naacl-long.111},
	booktitle = {Proceedings of the 2025 {Conference} of the {Nations} of the {Americas} {Chapter} of the {Association} for {Computational} {Linguistics}: {Human} {Language} {Technologies} ({Volume} 1: {Long} {Papers})},
	publisher = {Association for Computational Linguistics},
	author = {Gao, Mingqi and Hu, Xinyu and Lin, Li and Wan, Xiaojun},
	editor = {Chiruzzo, Luis and Ritter, Alan and Wang, Lu},
	month = apr,
	year = {2025},
	pages = {2199--2222}
}

@article{yeginbergen_dynamic_2025,
	title = {Dynamic {Knowledge} {Integration} for {Evidence}-{Driven} {Counter}-{Argument} {Generation} with {Large} {Language} {Models}},
	copyright = {Creative Commons Attribution 4.0 International},
	url = {https://arxiv.org/abs/2503.05328},
	doi = {10.48550/ARXIV.2503.05328},
	publisher = {arXiv},
	author = {Yeginbergen, Anar and Oronoz, Maite and Agerri, Rodrigo},
	year = {2025},
	note = {Version Number: 2}
}

@article{xiong_lllava-critic_2025,
	title = {{LLLaVA}-{Critic}: {Learning} to {Evaluate} {Multimodal} {Models}},
	copyright = {https://doi.org/10.15223/policy-029},
	shorttitle = {{LLLaVA}-{Critic}},
	url = {https://ieeexplore.ieee.org/document/11093772/},
	doi = {10.1109/CVPR52734.2025.01271},
	journal = {2025 IEEE/CVF Conference on Computer Vision and Pattern Recognition (CVPR)},
	publisher = {IEEE},
	author = {Xiong, Tianyi and Wang, Xiyao and Guo, Dong and Ye, Qinghao and Fan, Haoqi and Gu, Quanquan and Huang, Heng and Li, Chunyuan},
	month = jun,
	year = {2025},
	note = {Conference Name: 2025 IEEE/CVF Conference on Computer Vision and Pattern Recognition (CVPR)
ISBN: 9798331543648
Place: Nashville, TN, USA},
	pages = {13618--13628}
}

@inproceedings{kim_prometheus_2024,
	address = {Miami, Florida, USA},
	title = {Prometheus 2: {An} {Open} {Source} {Language} {Model} {Specialized} in {Evaluating} {Other} {Language} {Models}},
	shorttitle = {Prometheus 2},
	url = {https://aclanthology.org/2024.emnlp-main.248/},
	doi = {10.18653/v1/2024.emnlp-main.248},
	booktitle = {Proceedings of the 2024 {Conference} on {Empirical} {Methods} in {Natural} {Language} {Processing}},
	publisher = {Association for Computational Linguistics},
	author = {Kim, Seungone and Suk, Juyoung and Longpre, Shayne and Lin, Bill Yuchen and Shin, Jamin and Welleck, Sean and Neubig, Graham and Lee, Moontae and Lee, Kyungjae and Seo, Minjoon},
	editor = {Al-Onaizan, Yaser and Bansal, Mohit and Chen, Yun-Nung},
	month = nov,
	year = {2024},
	pages = {4334--4353},
}

@inproceedings{xu_instructscore_2023,
	address = {Singapore},
	title = {{INSTRUCTSCORE}: {Towards} {Explainable} {Text} {Generation} {Evaluation} with {Automatic} {Feedback}},
	shorttitle = {{INSTRUCTSCORE}},
	url = {https://aclanthology.org/2023.emnlp-main.365/},
	doi = {10.18653/v1/2023.emnlp-main.365},
	booktitle = {Proceedings of the 2023 {Conference} on {Empirical} {Methods} in {Natural} {Language} {Processing}},
	publisher = {Association for Computational Linguistics},
	author = {Xu, Wenda and Wang, Danqing and Pan, Liangming and Song, Zhenqiao and Freitag, Markus and Wang, William and Li, Lei},
	editor = {Bouamor, Houda and Pino, Juan and Bali, Kalika},
	month = dec,
	year = {2023},
	pages = {5967--5994}
}

@inproceedings{liu_x-eval_2024,
	address = {Mexico City, Mexico},
	title = {X-{Eval}: {Generalizable} {Multi}-aspect {Text} {Evaluation} via {Augmented} {Instruction} {Tuning} with {Auxiliary} {Evaluation} {Aspects}},
	shorttitle = {X-{Eval}},
	url = {https://aclanthology.org/2024.naacl-long.473/},
	doi = {10.18653/v1/2024.naacl-long.473},
	booktitle = {Proceedings of the 2024 {Conference} of the {North} {American} {Chapter} of the {Association} for {Computational} {Linguistics}: {Human} {Language} {Technologies} ({Volume} 1: {Long} {Papers})},
	publisher = {Association for Computational Linguistics},
	author = {Liu, Minqian and Shen, Ying and Xu, Zhiyang and Cao, Yixin and Cho, Eunah and Kumar, Vaibhav and Ghanadan, Reza and Huang, Lifu},
	editor = {Duh, Kevin and Gomez, Helena and Bethard, Steven},
	month = jun,
	year = {2024},
	pages = {8560--8579},
}

@inproceedings{perrella_guardians_2024,
	address = {Bangkok, Thailand},
	title = {Guardians of the {Machine} {Translation} {Meta}-{Evaluation}: {Sentinel} {Metrics} {Fall} {In}!},
	shorttitle = {Guardians of the {Machine} {Translation} {Meta}-{Evaluation}},
	url = {https://aclanthology.org/2024.acl-long.856/},
	doi = {10.18653/v1/2024.acl-long.856},
	urldate = {2026-05-18},
	booktitle = {Proceedings of the 62nd {Annual} {Meeting} of the {Association} for {Computational} {Linguistics} ({Volume} 1: {Long} {Papers})},
	publisher = {Association for Computational Linguistics},
	author = {Perrella, Stefano and Proietti, Lorenzo and Scirè, Alessandro and Barba, Edoardo and Navigli, Roberto},
	editor = {Ku, Lun-Wei and Martins, Andre and Srikumar, Vivek},
	month = aug,
	year = {2024},
	pages = {16216--16244}
}

@inproceedings{dinh_sciex_2024,
	address = {Miami, Florida, USA},
	title = {{SciEx}: {Benchmarking} {Large} {Language} {Models} on {Scientific} {Exams} with {Human} {Expert} {Grading} and {Automatic} {Grading}},
	shorttitle = {{SciEx}},
	url = {https://aclanthology.org/2024.emnlp-main.647/},
	doi = {10.18653/v1/2024.emnlp-main.647},
	booktitle = {Proceedings of the 2024 {Conference} on {Empirical} {Methods} in {Natural} {Language} {Processing}},
	publisher = {Association for Computational Linguistics},
	author = {Dinh, Tu Anh and Mullov, Carlos and Bärmann, Leonard and Li, Zhaolin and Liu, Danni and Reiß, Simon and Lee, Jueun and Lerzer, Nathan and Gao, Jianfeng and Peller-Konrad, Fabian and Röddiger, Tobias and Waibel, Alexander and Asfour, Tamim and Beigl, Michael and Stiefelhagen, Rainer and Dachsbacher, Carsten and Böhm, Klemens and Niehues, Jan},
	editor = {Al-Onaizan, Yaser and Bansal, Mohit and Chen, Yun-Nung},
	month = nov,
	year = {2024},
	pages = {11592--11610}
}

@misc{zhu_judgelm_2025,
	title = {{JudgeLM}: {Fine}-tuned {Large} {Language} {Models} are {Scalable} {Judges}},
	shorttitle = {{JudgeLM}},
	url = {http://arxiv.org/abs/2310.17631},
	doi = {10.48550/arXiv.2310.17631},
	publisher = {arXiv},
	author = {Zhu, Lianghui and Wang, Xinggang and Wang, Xinlong},
	month = mar,
	year = {2025},
	note = {arXiv:2310.17631 [cs]}
}

@inproceedings{ke_critiquellm_2024,
	address = {Bangkok, Thailand},
	title = {{CritiqueLLM}: {Towards} an {Informative} {Critique} {Generation} {Model} for {Evaluation} of {Large} {Language} {Model} {Generation}},
	shorttitle = {{CritiqueLLM}},
	url = {https://aclanthology.org/2024.acl-long.704/},
	doi = {10.18653/v1/2024.acl-long.704},
	booktitle = {Proceedings of the 62nd {Annual} {Meeting} of the {Association} for {Computational} {Linguistics} ({Volume} 1: {Long} {Papers})},
	publisher = {Association for Computational Linguistics},
	author = {Ke, Pei and Wen, Bosi and Feng, Andrew and Liu, Xiao and Lei, Xuanyu and Cheng, Jiale and Wang, Shengyuan and Zeng, Aohan and Dong, Yuxiao and Wang, Hongning and Tang, Jie and Huang, Minlie},
	editor = {Ku, Lun-Wei and Martins, Andre and Srikumar, Vivek},
	month = aug,
	year = {2024},
	pages = {13034--13054}
}

@inproceedings{lambert_rewardbench_2025,
	address = {Albuquerque, New Mexico},
	title = {{RewardBench}: {Evaluating} {Reward} {Models} for {Language} {Modeling}},
	isbn = {979-8-89176-195-7},
	shorttitle = {{RewardBench}},
	url = {https://aclanthology.org/2025.findings-naacl.96/},
	doi = {10.18653/v1/2025.findings-naacl.96},
	booktitle = {Findings of the {Association} for {Computational} {Linguistics}: {NAACL} 2025},
	publisher = {Association for Computational Linguistics},
	author = {Lambert, Nathan and Pyatkin, Valentina and Morrison, Jacob and Miranda, LJ and Lin, Bill Yuchen and Chandu, Khyathi and Dziri, Nouha and Kumar, Sachin and Zick, Tom and Choi, Yejin and Smith, Noah A. and Hajishirzi, Hannaneh},
	editor = {Chiruzzo, Luis and Ritter, Alan and Wang, Lu},
	month = apr,
	year = {2025},
	pages = {1755--1797}
}

@inproceedings{brabant_2026,
    title = {\'{E}valuation de métriques de RAG dans un contexte applicatif : une expérience, ses conclusions et ses limites},
    booktitle = {EvalLLM Workshop},
    author = {Brabant, Quentin},
    year = 2026
}
